# Fractional Local Neighborhood Intensity Pattern for Image Retrieval using Genetic Algorithm

[a] Shuvozit Ghose, [b]Abhirup Das, [c]Ayan Kumar Bhunia, [e]Partha Pratim Roy*

[a]Dept. of CSE, Institute of Engineering & Management, Kolkata, India Email- [a]shuvozit.ghose@gmail.com
[b]Dept. of CSE, Institute of Engineering & Management, Kolkata, India. Email- [b] abhirupdas.iem@gmail.com
[c]Dept. of ECE, Institute of Engineering & Management, Kolkata, India. Email- [c]2ayan.bhunia@gmail.com
[e]Dept. of CSE, Indian Institute of Technology Roorkee, India. Email- [e]proy.fcs@iitr.ac.in
[f*]email: proy.fcs@iitr.ac.in, TEL: +91-1332-284816*

## Abstract

In this paper, a new texture descriptor named "Fractional Local Neighborhood Intensity Pattern" (FLNIP) has been proposed for content based image retrieval (CBIR). It is an extension of an earlier work involving adjacent neighbors (local neighborhood intensity pattern). However, instead of considering two separate patterns for representing sign and magnitude information, one single pattern is generated. FLNIP calculates the relative intensity difference between a particular pixel and the center pixel of a 3×3 window by considering the relationship with adjacent neighbors. In this work, the fractional change in the local neighborhood involving the adjacent neighbors has been calculated first with respect to one of the eight neighbors of the center pixel of a 3×3 window. Next, the fractional change has been calculated with respect to the center itself. The two values of fractional change are next compared to generate a binary bit pattern. The descriptor is applied on four images- one being the raw image and the other three being filtered gaussian images obtained by applying gaussian filters of different standard deviations on the raw image to signify the importance of exploring texture information at different resolutions in an image. The four sets of distances obtained between the query and the target image are then combined with a genetic algorithm based approach to improve the retrieval performance by minimizing the distance between similar class images. The performance of the method has been tested for image retrieval on four databases and the proposed method has shown a significant improvement over many other existing methods.

*Keywords*- Local Neighborhood Intensity Pattern, Local Binary Pattern, Feature Extraction, Texture Feature.

# 1. Introduction

The advancement of technology has resulted in large volume of digital images available both offline and online which in turn has led to huge image databases. For retrieving information from this huge database based on a query image, many different types of retrieval methods have been proposed over the years. Some of the earlier methods used text based image retrieval. However, it suffers from some serious drawbacks. These are inappropriate metadata, lack of adequate information necessary for describing the image and very high time complexity. Thus, it has become less acceptable in the present context and led to the emergence of content based image retrieval.

**1.1 Motivation**

Rather than storing the huge image database, storing their extracted features in a feature database is more convenient. The features are used to represent and index the database. Feature extraction thus plays an important role in content based image retrieval (CBIR). Features can be of different types such as color, texture, shape and domain specific features such as human faces and fingerprint. Combining two or more types of features proves to be useful for content based image retrieval. The feature extraction method for CBIR is designed in such a way that it is able to retrieve information from images taken under different conditions with variations in illumination and intensity and capture the local microstructures present in an image as well as represent the local structural differences. A number of texture patterns have been developed for efficient image retrieval [1]–[8] in the recent past. However, most of these patterns have focused on comparing the center pixel of a 3×3 window of an image with only one of its neighbors at a time and ignored the remaining. Citing some drawbacks of this approach, some later works [3], [9] have highlighted the need of having separate sign and magnitude patterns to improve the discriminative power. However, they did not consider the mutual relationship among the neighboring pixels of the center which resulted in comparatively weak performance. The work done by Verma et al. [2], [10] stressed on this need which proved the importance of including the mutual relationship of adjacent neighbors of a pixel for encoding its relationship with the center. They considered the effect of two adjacent neighbors of a given neighboring pixel of the center. The work in [11] extended the idea using more adjacent neighbors and claimed a better performance. However, both [11] and [10] considered separate sign and magnitude patterns to encode each of them separately.

In this paper, we propose a new feature descriptor for image retrieval task. The striking advantage of our method over other methods is that we do not calculate separate sign and magnitude patterns. Instead, we focus on developing a descriptor that takes into account both the magnitude and sign information in a single descriptor. Thus, it reduces the feature dimension and computational complexity. Further, we explore the multi-resolution approach by applying the proposed descriptor on three different filtered versions of the same image obtained with gaussian filters of different standard deviations. This helps in encoding texture information present at different resolutions. The four sets of distances have been combined with genetic algorithm to improve the retrieval performance and make it more robust by reducing the dissimilarity between images of same class. The deep learning based approaches developed for image retrieval[12][13] have performed quite well for datasets of large size. However, on small sized datasets the performance of such networks have not been very good. In contrast, our network achieves good performance even on datasets having fewer images. Some other recent approaches have used pre-trained deep networks for image retrieval. The main problem of using such deep learning based networks is that they give misleading results on images considerably different from the images that they were trained on.

The remainder of the paper is organized in the following manner. The existing literature for content based image retrieval has been discussed in section 1.2. In section 1.3, we have discussed the main contributions of our work and its advantages with respect to existing techniques. In section 2, the proposed pattern have been discussed. Section 3 describes the proposed system framework. Section 4 describes the datasets used in our work and the results obtained on those datasets. The concluding part of the paper has been provided in the last section.

## 1.2 Related Work

Extensive research work has been done on content based image retrieval in the recent past. Wavelet transform based feature extraction has been performed in [14] . This technique suffers from serious drawbacks as it extracts information from an image in three different directions such as horizontal, vertical and diagonal. In order to overcome these drawbacks, a number of techniques have been proposed. For example, the Gabor wavelet feature based texture analysis was used in [15]. The concept of wavelet correlogram was used by Moghhadam et al. [16]. The concept of two dimensional rotated wavelet filters was used by Kokare et al. in [17]. The method which uses a combination of dual tree rotated complex wavelet filter and DT-CWT was proposed by the authors in [18]. Moreover, Nie et al. [53] proposed a performance prediction scheme for automatically predicting the performance of Web image search. They accomplish the task with query-adaptive graph-based learning based on the images ranking order and visual content. In another work, Nie et al. [54] exploited a heterogeneous probabilistic network to automatically estimate the relevance score of each image for image search with complex queries, which jointly couples three-layer relationships, spanning from semantic level to visual level. Moreover, to unravels the unreliable initial ranking list problem of the existing image reranking approaches, a heuristic approach is proposed to detect noun-phrase based visual concepts from complex query, instead of just treating individual terms as possible concepts.

A number of local patterns have been developed for CBIR over the years. The earliest texture descriptor that was used is Local Binary Pattern (LBP) [19]. It has also been used for a number of other tasks such as texture retrieval and classification in rotation invariant [19] and uniform versions [20]. The original LBP method generates an 8 bit binary string by considering pixels lying on the circumference of a circle of radius r and thresholding them with the center pixel. The main drawback of LBP is its inability to represent anisotropic structures due to circular sampling. This prompted the researchers to work on extending the operator as found in Dominant Local Binary Pattern (DLBP)[21], Center Symmetric Local Binary Pattern (CSLBP)[1], Block based local binary pattern (BLK LBP) [22], Completed Local Binary Pattern(CLBP)[23]. The Multi-Structure local binary pattern (Ms-LBP)[24] has been proposed to overcome the inefficiency of LBP in representing anisotropic structures. Another drawback of LBP is its sensitivity to noise. In order to overcome this drawback, Tan and Triggs proposed the local ternary pattern(LTP) [25] in which a ternary encoding is used to overcome the noise. Local Derivative pattern (LDP) [26] was invented by Zhang et al. for the purpose of face recognition. In this work, LDP has been shown to be the first order derivate of LBP. LDP was used in two separate forms, Local Edge Pattern for segmentation (LEPSEG) and Local Edge Pattern for image retrieval (LEPINV) [27] for different tasks. Peak Valley edge pattern (PVEP) [6] was obtained by calculating the first order derivative in four different directions. Local Maximum Edge Binary Pattern(LMEBP) [4] was developed by the researchers in for the purpose of image retrieval and object tracking. Murala et al. developed Local Tetra patterns (LTrP) [3] which represents the spatial structure of image texture in different directions and also calculated a magnitude pattern. The local Gabor XOR pattern was used for the purpose of face recognition in [28]. Noise-Resistant LBP (NR-LBP)[29] and a Robust Local Binary Pattern (RLBP) that contains both sign and magnitude information was also proposed in [30] to reduce noise of LBP feature. Gaussian filters and RGB color space was used by Murala et al in [31]. In another work, the authors extracted the LTP feature from different directions. This was named as Spherical Symmetric 3D Local

Ternary Patterns [32]. Content based image retrieval also has tremendous applications for biomedical image retrieval where it has proved to be useful in handling a large image database. Although a number of local pattern based texture features have been developed for image retrieval in the recent past, not many have considered separate magnitude and sign patterns. The directional binary wavelet patterns were developed by Murala et al. for biomedical image retrieval and indexing [33]. Besides all these uses, there are other implementations which used the concept of LBP and have made further improvements on it. Among these, Local Bit-plane Decoded Pattern(LBDP)[34], Local Mesh Pattern (LMP) [35], Average Local Binary Pattern(ALBP) [36], Ellipse Topology [37], three and four patch LBP[38] etc. are worth mentioning. In Directional Local Extrema Pattern [39], the feature was mainly obtained using the edge details. In this feature extraction technique by edge information, four directions of the image were utilized and further improvement on it was done in [40][4] which is used for retrieving images. The authors proposed a novel technique for texture image retrieval based on tetrolet transforms [41]. A work on query by saliency content retrieval (QSCR) image based on human visual attention models has been proposed by Papushoy et al. in [42].

The concept of Gray Level Co-occurrence matrix (GLCM) was used by Haralick for classification of the image where some statistical features are used to consider the intensity of the pixels of images and mainly focuses on the occurrences of a particular kind of pattern. In many cases, Prewitt operator is operated on the image and then GLCM is applied to calculate statistical parameters, rather than directly applying GLCM operator on the real image[43]. Again GLCM is implemented for extracting the feature vector in Center Symmetric Local Binary Co-occurrence Pattern [1], where the feature is extracted based on the diagonally symmetric elements about the center. In Local Neighborhood Difference Pattern(LNDP) [2], only the nearest element to a particular pixel is considered for the feature extraction and the ultimate feature extraction for image retrieval is performed by concatenating this LNDP feature with LBP feature.

### 1.3 Local Binary Patterns

Ojala et al. [44] originally proposed local binary pattern for texture classification. Due to ease of computation, its use was extended to object tracking[45], facial expression recognition [46] and medical imaging [47]. It has also been popularly used for the purpose of image classification. This method considers a small window of an image and the intensity difference between the center pixel and its K neighbors lying on the circumference of a circle of radius

R is encoded in binary form. Considering the center pixel as the threshold, a binary value (0 or 1) is assigned to each of the surrounding pixels (as given in eqn. 2). These binary bits are then multiplied by some specific weights which sum up to produce a decimal value which replaces the center pixel. This has been explained diagrammatically in Fig. 1. Thus, a local binary map of the image is generated by replacing each center pixel with its binary pattern value in a similar manner. The feature vector is calculated by creating a histogram of this local binary map. The formula for local binary pattern and the histogram is defined in eqn. (1)-(4).

$$LBP_{K,R} = \sum_{m=1}^{K} 2^{m-1} \times \emptyset_1(I_m, I_c) \quad (1)$$

$$\emptyset_1(I_m, I_c) = \begin{cases} 1, & I_m \geq I_c \\ 0, & \text{otherwise} \end{cases} \quad (2)$$

$$H(L) = \sum_{m=1}^{X} \sum_{n=1}^{Y} \emptyset_2(LBP_{K,R}(m,n), L); \quad (3)$$

$$\emptyset_2(c_1, c_2) = \begin{cases} 1, & c_1 = c_2 \\ 0, & \text{otherwise} \end{cases} \quad (4)$$

Here, K and R represent the number of neighboring pixels and radius respectively and the maximal value of LBP pattern is $I_m$ denotes the $m^{th}$ surrounding pixel and $I_c$ denotes center pixel. Eqn. (3) is used to compute the final histogram of the pattern map. In eqn. (3), L ∈ [0, V], where V is the maximum value of LBP pattern. An example of LBP calculation is shown in Fig. 1. In the next subsection, our proposed pattern has been explained in brief.

### 1.4 Main Contributions

The main contributions of our work may be stated in brief as: First, we develop a novel texture descriptor for content based image retrieval named FLNIP which is based on comparing the fractional change in intensity in the adjacent neighborhood between the center pixel of a 3×3 window and its neighboring pixel ($I_j$) where j= {1,2,3….8}. Second, we perform a multi-resolution analysis of the proposed pattern using gaussian filters of different standard deviations to explore the texture information at different scales. Third, the distances obtained by applying the descriptor on raw and filtered images have been optimally combined by using genetic algorithm.

**1.5 Difference with Local Neighborhood Intensity Pattern**

In our previous work on local neighborhood intensity pattern [11], we introduced the concept of local neighbors and generated a sign and magnitude pattern by considering the relative intensity difference between a particular pixel and the center pixel by considering its adjacent neighbors. In this work, we further extend the concept of exploiting texture information present in the local neighborhood of a particular pixel by considering the fractional change in the intensity in the local neighborhood and the center pixel of a 3×3 window with respect to a particular pixel and then comparing those values. The advantage that it offers over LNIP (Local Neighborhood Intensity Pattern) is that we do not require separate sign and magnitude patterns to encode neighborhood information. However, since the proposed pattern encodes the fractional change, the pattern becomes more informative about the intensity change both in terms of sign and magnitude. The difference arises due to the fact that rather than comparing the changes in magnitude directly, we compare the fractional change. Hence, it is able to overcome the drawback of LNIP and encode neighborhood information more effectively.

## 2. Proposed Pattern

In this section, we discuss in detail the proposed pattern and the two different approaches adapted to make the pattern superior to other local patterns developed so far for content based image retrieval.

**2.1 Fractional Local Neighborhood Intensity Pattern**

The descriptor named fractional local neighborhood intensity pattern is based on comparing the intensities of the center pixel of a 3×3 window with its neighboring pixels. It is inspired from the adjacent neighbor concept in LNIP [11]. In LNIP, it was shown that adjacent neighbors of a particular pixel contain useful texture information and it must be taken into account for calculating its relative intensity difference with respect to the center pixel of a 3x3 window rather than performing a direct comparison as used in LBP (Local Binary Pattern). It first calculates the sum of the absolute difference of pixel intensities of the adjacent neighbors of a particular neighbor ($I_k$) of the center pixel, where k={1,2,3,4,….,8} . In eqn. 5 and 6, $\alpha_k$ denotes the set of adjacent neighbors of a given pixel $I_k$ for k={1,3,5,7}. The variable M indicates the number of adjacent neighbors. For k={1,3,5,7}, M=4 and for K={2,4,6,8}, M=2. The term $\mu_k$ in eqn. 7 represents the average sum of intensity differences of the adjacent neighbors of $I_k$ divided by $I_k$ for M neighbors. We call it 'fractional change' with respect to ($I_k$) in this work. Following the same set of steps, we calculate the fractional change with respect to the center pixel as $\mu_c$ as given in eqn. 8. If the 'fractional change' with respect to the particular neighbor ($\mu_k$) is more than the fractional change with respect to the center($\mu_c$), it is encoded as 1, otherwise it is encoded as 0. Sign(a,b) represents the sign operation in eqn. 10 which has been used to calculate the thresholding relation given in eqn. 9. This is repeated for all the remaining neighbors of the center pixel. Thus, we generate an 8 bit binary pattern of 0's and 1's. The decimal value for the pattern is given by eqn. 11. This gives a feature descriptor having a dimensionality of 256. The diagram shown in Fig. 2 shows the pattern calculation procedure for $I_1$. The process can be repeated to obtain bit pattern for remaining positions. Fig.2 shows an example of calculation of the bit pattern using fractional local neighborhood intensity pattern. To avoid complexity, only the calculation of the bit value for $I_1$ has been shown. The fractional change with respect to $I_c$ in our example is 2.53 while that for $I_1$ is 2.31. Since 2.53 is greater than 2.31, the bit for $I_1$ is 0. Please note, in order to avoid division by zero, we map [0,255] to [1,256] for pixel values.

$$\alpha_k = \{I_{1+mod(k+5,7)}, I_{1+mod(k+6,9)}, I_{k+1}, I_{mod(k+2,8)}\} \ \forall \ k = 1,3,5,7 \quad (5)$$

$$\alpha_k = \{I_{k-1}, I_{mod(k+1,8)}\} \ \forall \ k = 2,4,6,8 \quad (6)$$

$$\mu_k = \frac{1}{M}\sum_{m=1}^{M} \frac{|\alpha_k(m) - I_k|}{I_k} \quad (7)$$

$$\mu_c = \frac{1}{M}\sum_{m=1}^{M} \frac{|\alpha_k(m) - I_c|}{I_c} \quad (8)$$

$$\mu(I_k, I_c) = Sign(\mu_k, \mu_c) \quad (9)$$

$$Sign(a, b) = \begin{cases} 1, & a \geq b \\ 0, & otherwise \end{cases} \quad (10)$$

$$FLNIP(I_c) = \sum_{k=1}^{8} 2^{k-1} \times \mu(I_k, I_c) \quad (11)$$

| $I_6$ | $I_7$ | $I_8$ |
|---|---|---|
| $I_5$ | $I_c$ | $I_1$ |
| $I_4$ | $I_3$ | $I_2$ |

(i)

| 85 | 30 | 39 |
|---|---|---|
| 10 | 42 | 55 |
| 54 | 27 | 38 |

(ii)

| 1 | 0 | 0 |
|---|---|---|
| 0 |   | 1 |
| 1 | 0 | 0 |

(iii)

| 8 | 4 | 2 |
|---|---|---|
| 16 |   | 1 |
| 32 | 64 | 128 |

(iv)

| 8 | 0 | 0 |
|---|---|---|
| 0 | 41 | 1 |
| 32 | 0 | 0 |

(v)

**Fig.1.** An example to demonstrate the calculation of Local Binary Pattern (LBP) (i) a 3×3 window with general notations of the center and its neighboring elements (ii) an example of a 3×3 window (iii)-(v) threshold neighboring pixels with center and assigning binary values accordingly.

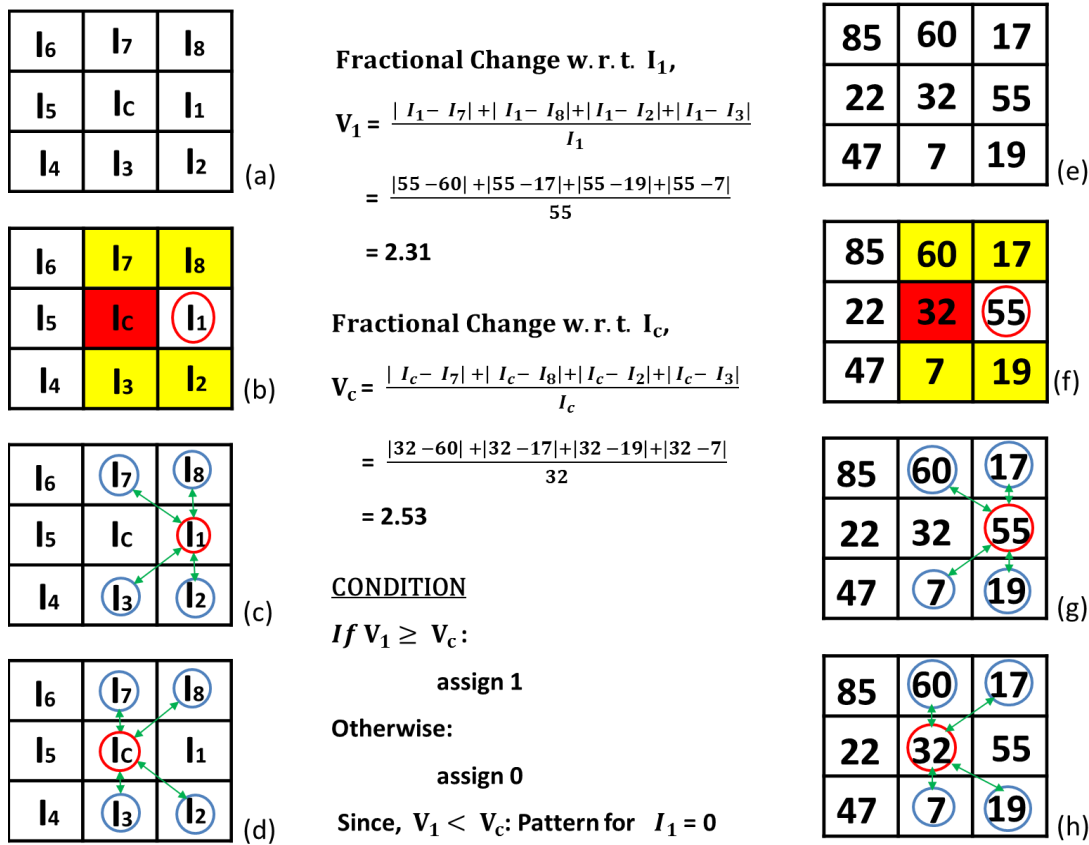

Fractional Change w.r.t. $I_1$,

$$V_1 = \frac{|I_1 - I_7| + |I_1 - I_8| + |I_1 - I_2| + |I_1 - I_3|}{I_1}$$

$$= \frac{|55 - 60| + |55 - 17| + |55 - 19| + |55 - 7|}{55}$$

$$= 2.31$$

Fractional Change w.r.t. $I_c$,

$$V_c = \frac{|I_c - I_7| + |I_c - I_8| + |I_c - I_2| + |I_c - I_3|}{I_c}$$

$$= \frac{|32 - 60| + |32 - 17| + |32 - 19| + |32 - 7|}{32}$$

$$= 2.53$$

**CONDITION**

If $V_1 \geq V_c$:
  assign 1
Otherwise:
  assign 0

Since, $V_1 < V_c$: Pattern for $I_1 = 0$

**Fig.2.** Example showing the calculation of the proposed Fractional Local Neighborhood Intensity Pattern (FLNIP) for a 3×3 window with general notations of the center and its neighboring elements. Diagrams (b) through (h) illustrate the calculation of the proposed pattern for the neighbor $I_1$ of the center pixel.

## 2.2. Gaussian filters for multi-resolution analysis

All the different textures existing in an image may not be prominent at a given resolution. A particular texture may be more prominent over others at a given resolution. This motivates us to increase the robustness of the proposed descriptor by applying a set of different isotropic gaussian filters for performing multi-resolution analysis. The filters are designed choosing three different values of standard deviation. The gaussian filter may be expressed as:

$$G(x, y, \sigma) = \frac{1}{2\pi\sigma^2} e^{-(x^2 + y^2)/2\sigma^2} \quad (12)$$

The filtered image can be represented as:

$$H(x, y) = G(x, y, \sigma) * I(x, y) \quad (13)$$

In eqn. 12, x and y represent the spatial coordinates of the image. The standard deviation is represented by $\sigma$. The convolution operation in eqn.13 is represented by $*$. In this paper, we consider three different scales to obtain three filtered images. The three different values of standard deviation considered for gaussian filtering are 0.5, 0.8 and 1. These values are selected based on experimental results. We obtain three different histograms for filtered images. On performing feature extraction after filtering, four sets of distance values between query and database images (three sets of distance for filtered images and one set of distance values for raw image) are combined in a weighted manner using genetic algorithm to produce a better retrieval accuracy.

**2.3. Genetic Algorithm for distance combination**

In this paper, we propose to combine the different distances (between the query and the database images) obtained by applying the descriptor on filtered images of different standard deviations and the raw image itself using genetic algorithm. In our problem we aim to improve the retrieval performance by maximizing precision with the help of genetic algorithm. For this, the intra class similarity must be made higher and inter class similarity must be made lower. To begin with, we randomly initialize a set of possible solutions which constitutes a population. Each individual of the population is called a chromosome. For our case, a chromosome is a set of four weights learned with the algorithm. Each chromosome consists of genes, which in this case are the individual weights assigned to each distance. Each chromosome in our approach has four genes. For our approach, we have used roulette wheel selection with two point crossover and a mutation rate of 0.01(1%). In eqn.14 $d(Q, F)$ is the distance between a query image feature and the feature of the database image where Q is the feature vector representation of the query image and $F_i$ is the feature vector of the database image. The length of the feature vector is p. As mentioned earlier, we have to combine four distances using genetic algorithm. This implies that four weights will have to be learned. Table 1 shows the different parameters used in our algorithm implementation.

$$d(Q, F) = \sum_{j=1}^{4} W_j \left( \sum_{i=1}^{p} \left| \frac{Q - F_i}{1 + Q + F_i} \right| \right) \quad (14)$$

The precision in eqn. 22 refers to the precision for each category. Eqn. 24 represents the average precision of all the categories and serves as the fitness function for our problem. After obtaining the final set of distances by using the weights learned, the k images having least value of distance are treated as similar class images, where k refers to the number of images in each category for a particular dataset.

## 3. Overall Framework

A flow diagram of the proposed method is presented in Fig. 3 and algorithm for the same has been described below. The proposed framework has been divided into two sections. The first part describes the system framework. Here, an image is given as the input. Different sets of feature representations are obtained after application of FLNIP on raw and filtered images. In part 2, a query image is given as input. Retrieval of image begins by applying FLNIP on raw and filtered versions of the query image. The different distances between query and retrieved images obtained after applying FLNIP on the raw and filtered versions of the query image are combined by performing a weighted combination using genetic algorithm. The images are retrieved based on similarity measure.

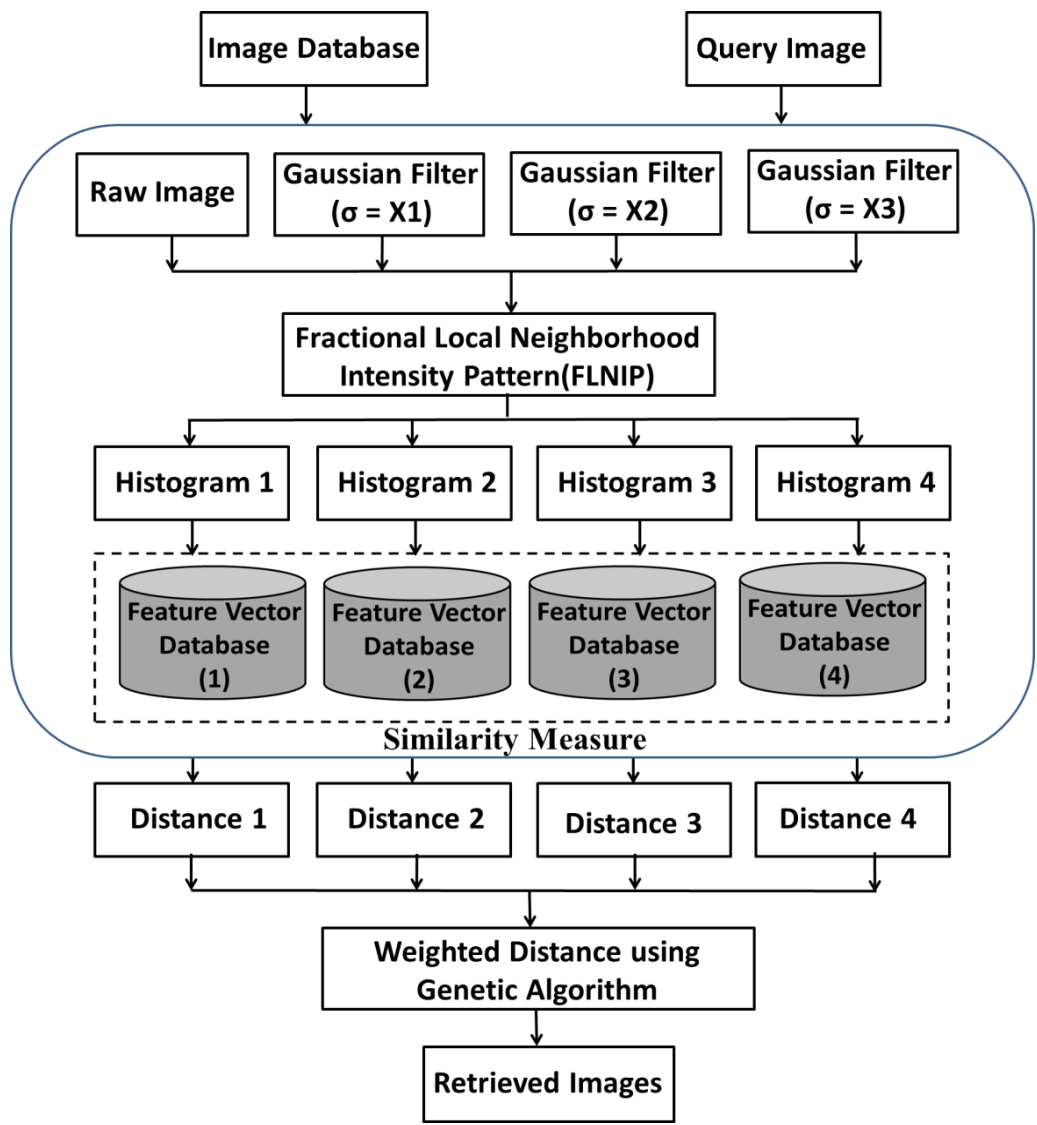

Fig. 3. Proposed system framework.

## 3.1 System Framework Algorithm

**Section 1: Construction of feature vector**

For the first part of our problem, we sample an image from the image database and convert it into gray scale image if the image is colored. To perform the multi-resolution analysis, application of gaussian filters of different standard deviations on the original image is performed. The final step is the application of the proposed descriptor on raw and filtered images of different standard deviations followed by the computation of the respective histograms.

**Section 2: Image retrieval using the proposed method**

For the second part of our problem, we consider each and every image as a query image and repeat the multi-resolution approach prior to descriptor application. This helps us obtain four sets of distances between query and each database image. For the next step, we combine the distances by using a weighted approach using genetic algorithm to obtain an optimal distance. The performance is evaluated using the metrics like precision and recall.

## 3.2. Similarity Measure

For retrieving the images in content based image retrieval system, apart from texture feature representation, similarity measure plays a pivotal role. The similarity measure or metric is used to calculate the distance between the query image and every image from the database in the feature space and then rank the images accordingly. Based on this ranking, indexing is done and indices with lower measures are sorted out as the set of retrieved images. A set of five distance measures have been used in our work for calculation of similarity matching as represented in eqn.15-19.

a. d1 distance:

$$\tau_{D,q_m} = \sum_{n=1}^{l} \left| \frac{\cap_d^m(n) - \cap_{q_m}(n)}{1 + \cap_d^m(n) + \cap_{q_m}(n)} \right| \quad (15)$$

b. Euclidean Distance:

$$\tau_{D,q_m} = \left(\sum_{n=1}^{l} |(\cap_d^m(n) - \cap_{q_m}(n))^2|\right)^{1/2} \quad (16)$$

c. Manhattan Distance:

$$\tau_{D,q_m} = \sum_{n=1}^{l} |\cap_d^m(n) - \cap_{q_m}(n)| \quad (17)$$

d. Canberra Distance:

$$\tau_{D,q_m} = \sum_{n=1}^{l} \left|\frac{\cap_d^m(n) - \cap_{q_m}(n)}{\cap_d^m(n) + \cap_{q_m}(n)}\right| \quad (18)$$

e. Chi-square Distance:

$$\tau_{D,q_m} = \frac{1}{2}\sum_{n=1}^{l} \frac{(\cap_d^m(n) - \cap_{q_m}(n))^2}{\cap_d^m(n) + \cap_{q_m}(n)} \quad (19)$$

Here, $\tau_{D,q_m}$ represents the distance function for database D and query image $q_m$, $l$ represents the length of the feature vector. $\cap_d^m(n)$ and $\cap_{q_m}(n)$ are feature vector of $m^{th}$ database image and query image respectively.

## 4. Experimental Results and Analysis:

The suitability of the proposed descriptor for CBIR (Content Based Image Retrieval) is verified by the results obtained on different datasets for image retrieval including texture datasets like Stex and Brodatz texture database. One biomedical image and one face dataset have also been considered for evaluation, namely the OASIS database and ORL face database. For each dataset, we have evaluated a number of different parameters including precision, recall, F-score and average retrieval rate (ARR). The precision is defined as the ratio of the true positive rate to the sum of the true positive and false positive rate. For an image retrieval system, it is a ratio of the total number of correct images retrieved from the database to the total number of images retrieved from the database. Correct images implies the number of images relevant for a given query image. On the other hand, recall is a function of the true positive images and false negative images in the retrieval system. For any image retrieval system, precision and recall generally share an inverse relationship. In other words, precision generally decreases and recall generally increases with increase in the number of images retrieved. Precision may be computed as given in eqn.20 and recall may be computed as given in eqn. 21:

$$\text{Precision Rate}(q, P_i) = \frac{\text{True Positive Rate (TP)}}{\text{True Poasitive Rate(TP)} + \text{False Positive Rate(FP)}} \quad (20)$$

where, q represents the query image and $P_i$ represents the precision value for the $i^{th}$ query image.

Similarly, the eqn. for recall may be represented as:

$$\text{Recall Rate}(q, N_i) = \frac{\text{True Positive Rate (TP)}}{\text{True Positive Rate(TP)} + \text{False Negative Rate(FN)}} \quad (21)$$

Here, $N_i$ indicates the number of correct images in the database, i.e., number of images in each category of the dataset.

The average precision rate may be calculated as follows:

$$P_{avg}(K) = \frac{1}{m}\sum_{i=1}^{m} P_i \quad (22)$$

In eqn. 22, $P_{avg}(K)$ represents the average precision rate for category (K), where m is the total no of images in that category. Similarly, the recall rate for each category may be expressed as given in eqn. 23

$$R_{avg}(K) = \frac{1}{m}\sum_{i=1}^{m} R_i \quad (23)$$

The total precision rate and total recall rate can be computed for our experiment using eqn. 24 and 25 where K is the total no of categories in the database.

$$P_{tot}(K) = \frac{1}{K}\sum_{c=1}^{K} P_{avg}(K) \quad (24)$$

$$R_{tot}(K) = \frac{1}{K}\sum_{c=1}^{K} R_{avg}(K) \quad (25)$$

For our experiments, we have also determined the F-score as given in eqn.26. The F-score is mathematically expressed as the harmonic mean of the precision and recall. We can write the F-score mathematically as:

$$F-score = \frac{2\times Precision \times Recall}{Precision + Recall} \quad (26)$$

The performance of the method has been shown for all datasets by measuring the precision, recall and F-score and comparing with state-of-the-art-methods. The feature vector length and extraction time has been determined for the proposed method and compared with some existing techniques for image retrieval in Table 4. The proposed pattern has been compared with six other image retrieval techniques like center symmetric local binary pattern, local binary pattern, center symmetric local binary co-occurrence pattern, local tri-directional patterns, local edge pattern for segmentation. We follow the same experimental setup like LNIP[11].

### 4.1 Proposed experimental approach for comparison with other methods

We have performed the experiments using three different approaches. First one is simple feature extraction using FLNIP from query image and database images and ranking based on similarity. The second is the linear combination of different distances obtained between query and database image using raw and filtered versions of the image. The third approach is a weighted combination of the same set of distances using genetic algorithm based optimization. The third approach clearly performs better than the rest. Our distance weighting strategy using G.A. also performs better than recently used techniques like adaptive weighted fusion[48] and the one used in [49] for automatic weight selection. The distance metric plays a pivotal role in determining the suitability of the proposed method. For our work, we have applied d1 distance metric for evaluating the performance of all descriptors. It is important to note that the proposed method has been compared with other descriptors without applying the multi-resolution and distance combination techniques on them.

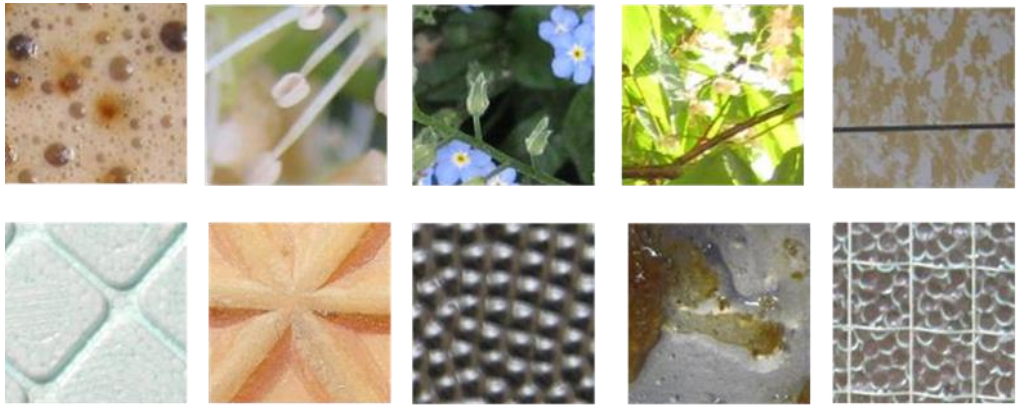

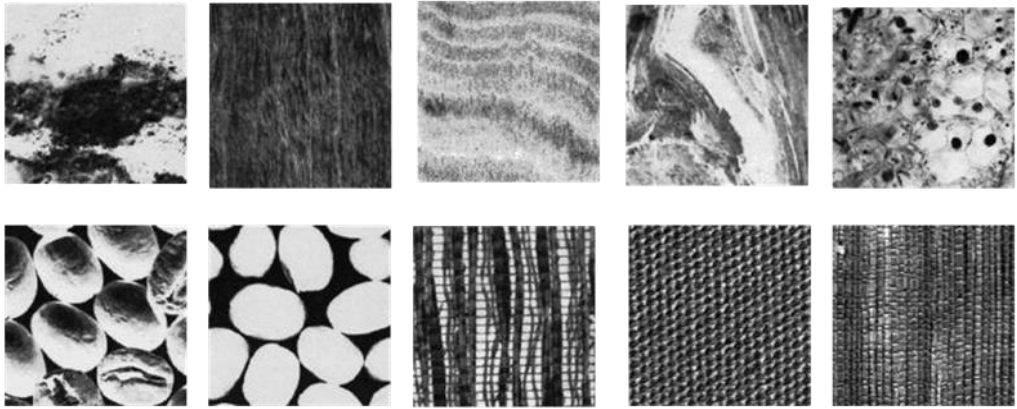

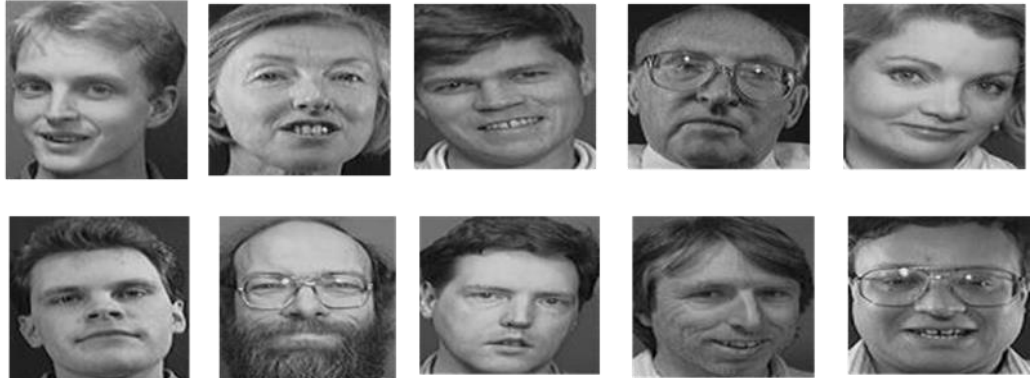

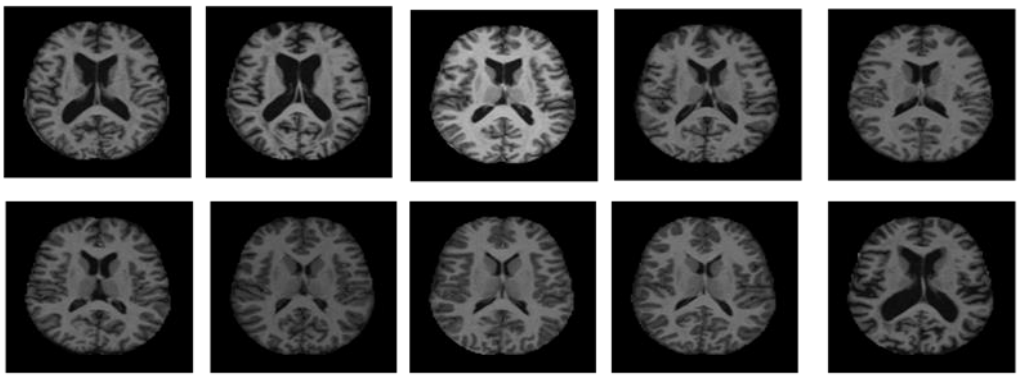

**Fig. 4. Sample images from different datasets.**

**Database 1:**

For the first experiment, we have used the Brodatz[50] texture database. The database consists of images of size 640×640 divided into sub images of size 128×128. Thus, there are 25 sub images of each image. There are a total of 112 categories of images in this dataset. The total number of images for this dataset is 2800. Keeping consistency with the other datasets, the same metrics have been used to evaluate the performance of the proposed approach for this dataset. We have plotted the precision, recall and F-score curves for this dataset as shown in Fig. 5. Some sample images from this database have been shown in Fig 4(b). Fig. 6 shows images retrieved for the given query image.

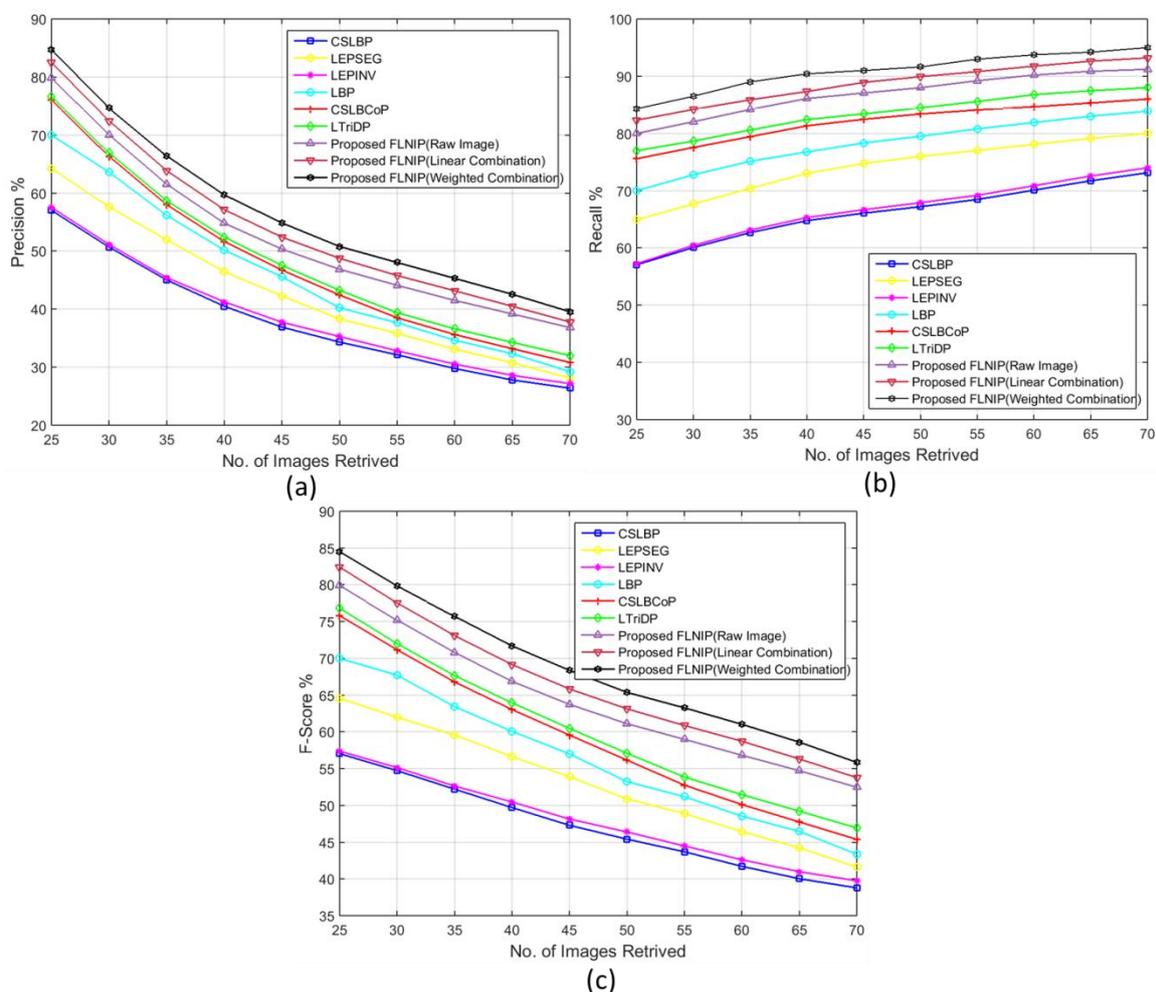

h

**Fig. 5. Precision, recall and F-score with number of images retrieved for Brodatz database**

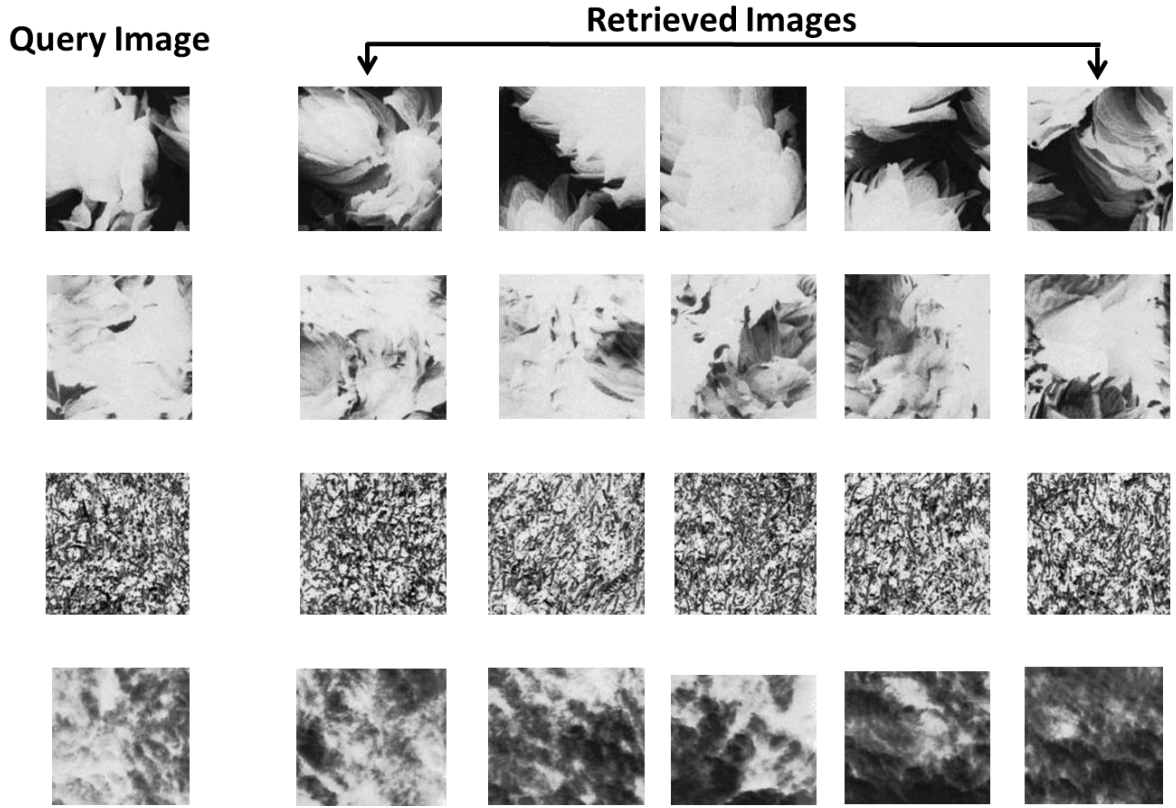

Fig. 6. Query image and retrieved images from Brodatz database

**Database 2:**

We have evaluated the performance of our method on The Open Access Series of Imaging Studies (OASIS)[51] dataset. This is a biomedical image dataset which is widely used to verify the performance of a biomedical image retrieval system. The dataset consists of a collection of 421 subjects with ages between 18 and 96 years. There are a total of 124, 102, 89 and 106 images in each category. The grouping has been done based on the shape of the ventricular. Fig. 4(d) represents some sample images from this database. Precision with varying number of retrieved images and precision for different groups of images for this dataset are shown in Fig. 7. The query and retrieved images have been shown in Fig. 8. Some details regarding the dataset are given in Table 2. The Average Retrieval Rate (ARR) has been calculated for this dataset similar to other datasets. Table 3 shows the results. The results obtained with different distance metrics are shown in Table 5.

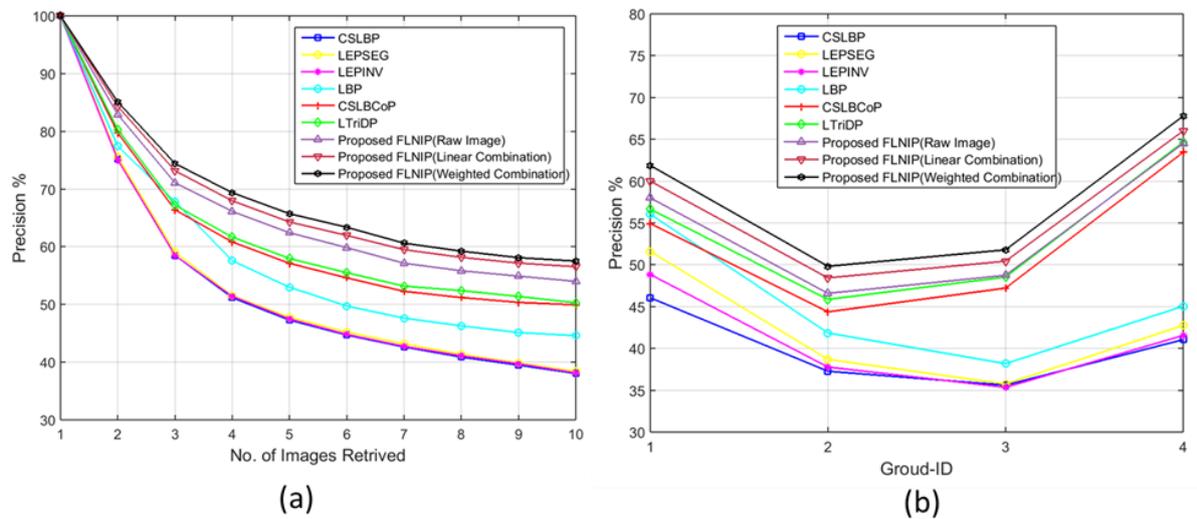

**Fig.7. Precision with number of images retrieved and precision with Group-ID for OASIS database.**

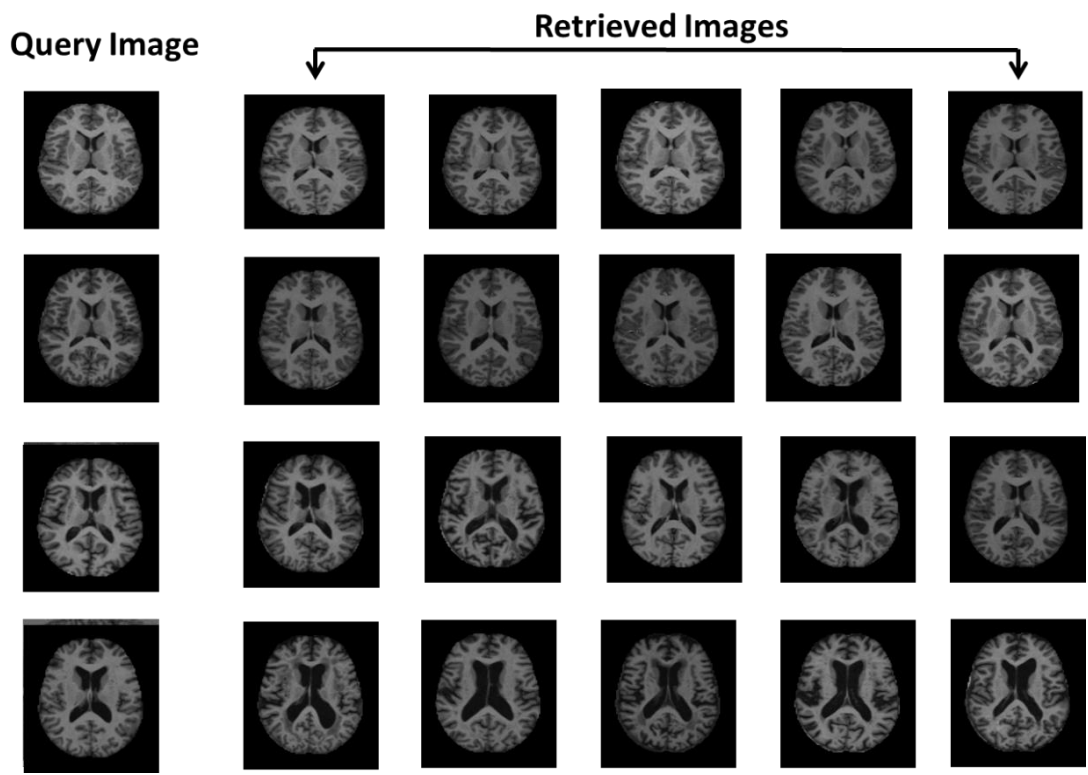

**Fig. 8 Query image and retrieved images from OASIS database.**

Table 2: Details of MRI Data Acquisition.

| Sequence | MP-RAGE |
|---|---|
| TR (msec) | 9.7 |
| TI (msec) | 20 |
| Flip angle (o) | 10 |
| TE (msec) | 4.0 |
| Resolution (pixels) | 176,208 |
| Orientation | Sagittal |
| TD (msec) | 200 |
| Thickness, gap (mm) | 1.25, 0 |

**Database 3:**

The performance of the proposed method has been evaluated on the ORL[52] database of facial images for face image retrieval. There are a total of 40 users in this dataset and each user has 10 images. The database has been created by AT&T laboratories, Cambridge. Thus, there are a total of 400 images in this dataset. For some users, the images were taken at different times and under varying illumination and expression. Sample images from this database are shown in Fig. 4(c). The results of precision, recall and F-score on various methods are shown in Fig 9. Query and retrieved images corresponding to them are shown in Fig.10. The Average Retrieval Rate (ARR) has been calculated for this dataset similar to other datasets. Table 3 shows the results. The results obtained with different distance metrics are shown in Table 5 while Table 4 shows the time required for feature extraction for our framework and those for other methods.

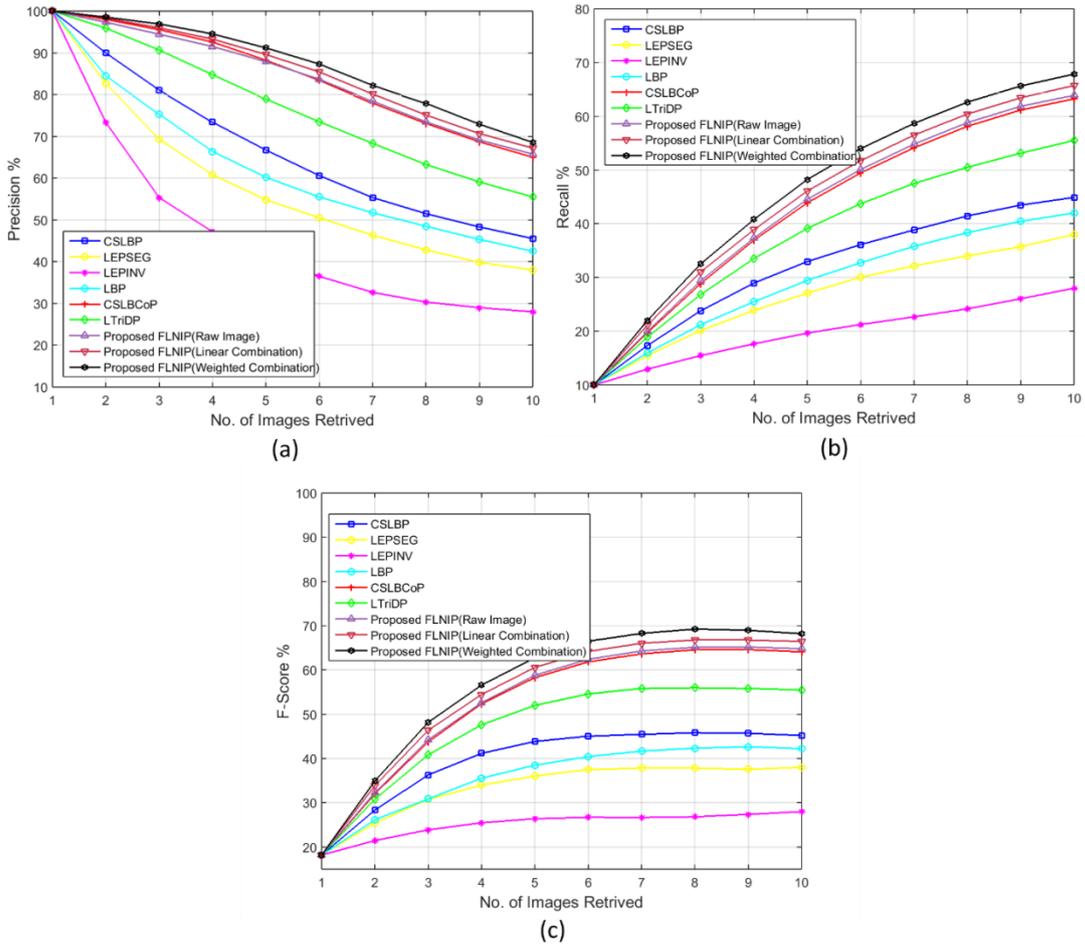

**Fig. 9. Precision, recall and F-score with number of images retrieved for the ORL database.**

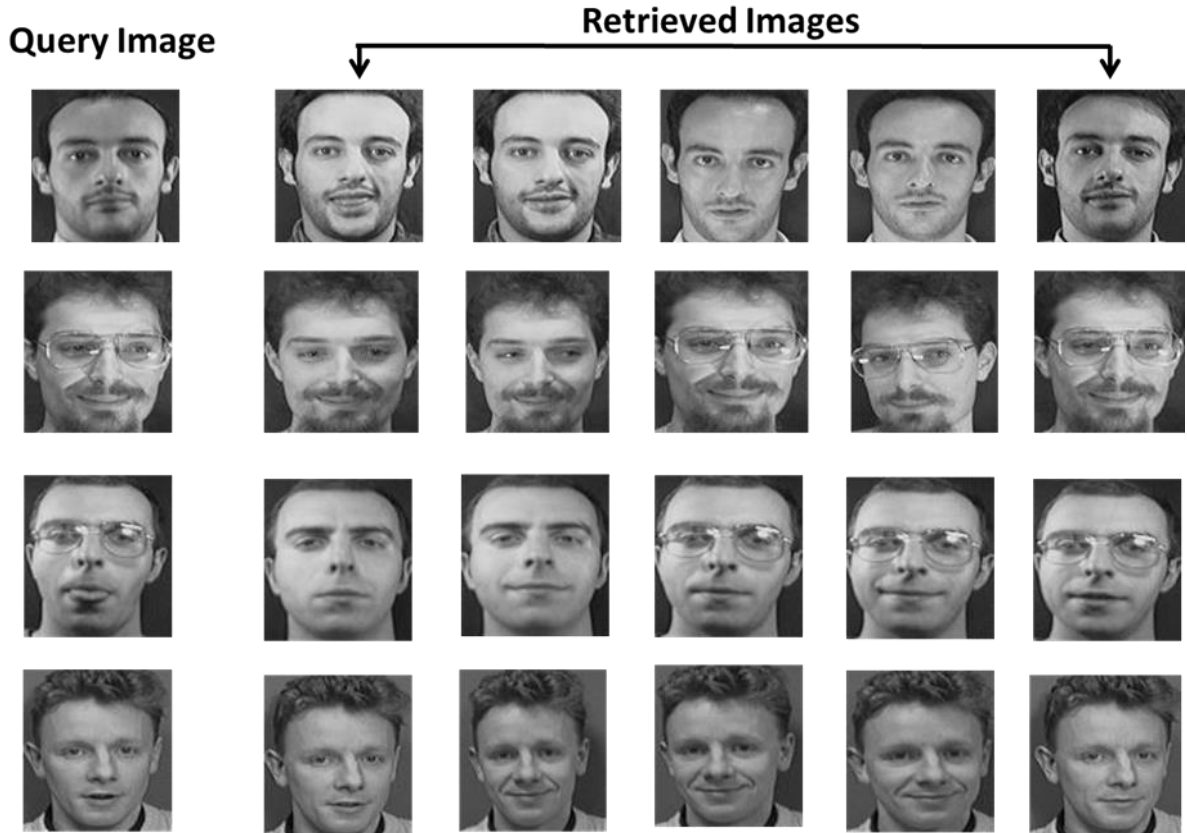
**Fig. 10 Query image and retrieved images from ORL database.**

**Database 4:**

The fourth database used for the experimental study in our work is the Salzburg texture database. The Salzburg[1] texture Database contains images of size 128×128. There are a total of 7616 images. There are a total of 476 categories and each category contains 16 images. Different types of textures like wood, rubber, etc. are presented in the database. Sample images from the database are presented in Fig. 4(a). Each image of the database is treated as a query image. Fig. 12 shows some query images and their corresponding retrieved images. The number of images retrieved for each category for this experiment is initially considered as 16. This is increased in small steps of 16 images. The maximum number of images for this dataset that have been retrieved in our experiment is 112. The precision, recall and F-score have been plotted in Fig.11.

---

[1] http://www.wavelab.at/sources/STex/

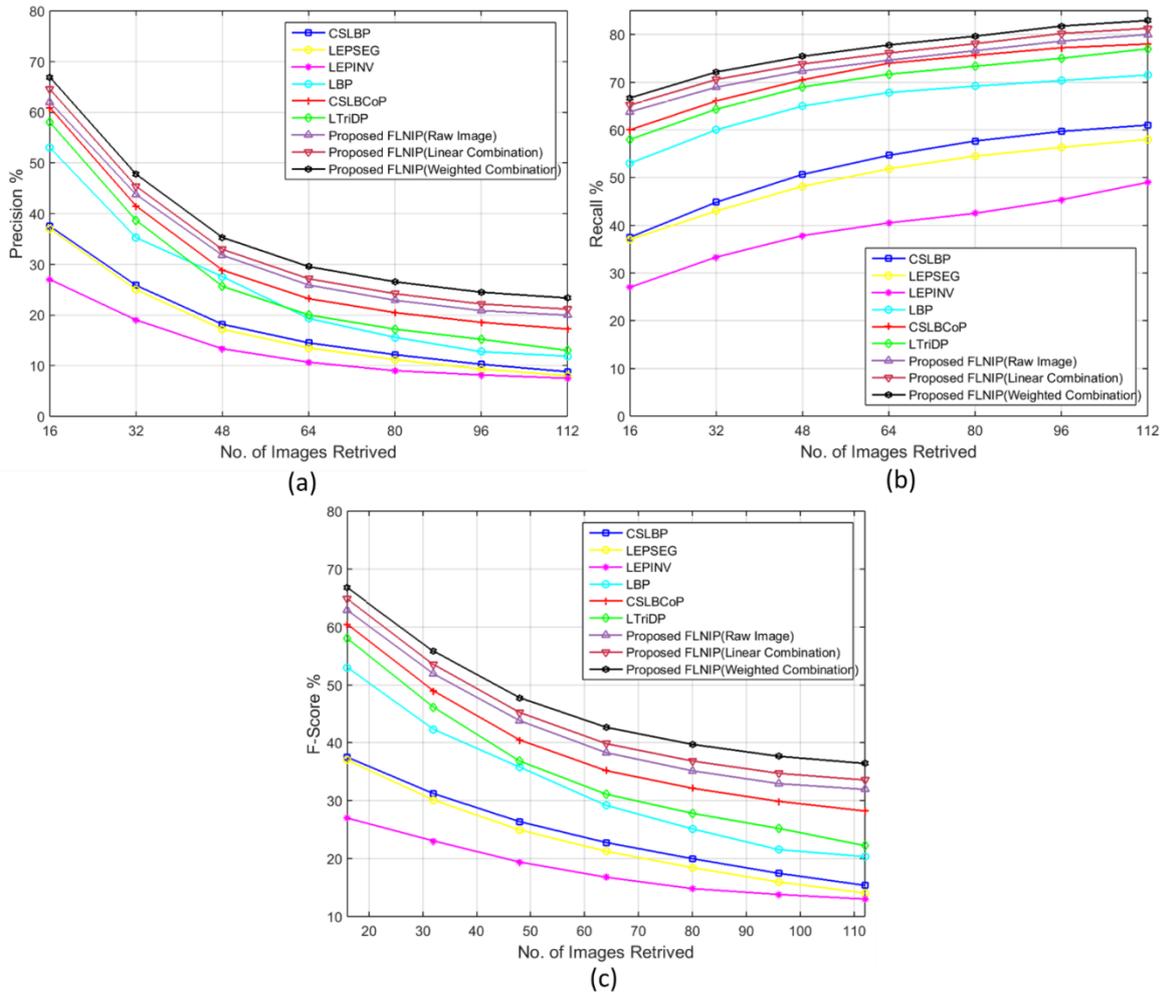

**Fig 11. Precision, recall and F-score with number of images retrieved for Salzburg texture database**

## 4.2 Analysis of experimental results obtained with the proposed approach

In this section, we present a detailed analysis of the results obtained with our experiments. The graphs of precision, recall and F-score clearly shows the superiority of the proposed texture feature compared to recent state-of-the-art texture descriptors. The results obtained on performing image retrieval using the proposed texture pattern have presented with the help of three graphs pertaining to three sets of experiments. The first graph shows the performance of retrieval using the proposed multi-resolution approach. The second graph shows the performance of retrieval by linearly combining the four sets of distances between the actual and retrieved images. The third graph shows the performance improvement on combining the distances using genetic algorithm based learning. The performance has been evaluated by varying the number of images retrieved. For brodatz, we have initially retrieved 25 images and then increased the number of images retrieved in steps of 5 images. Thus, we have retrieved a maximum of 70 images for brodatz. On the other hand, for Stex database, the number of images retrieved have been varied from 16 to 112 while for the face database, it has been varied from 1 to 10.

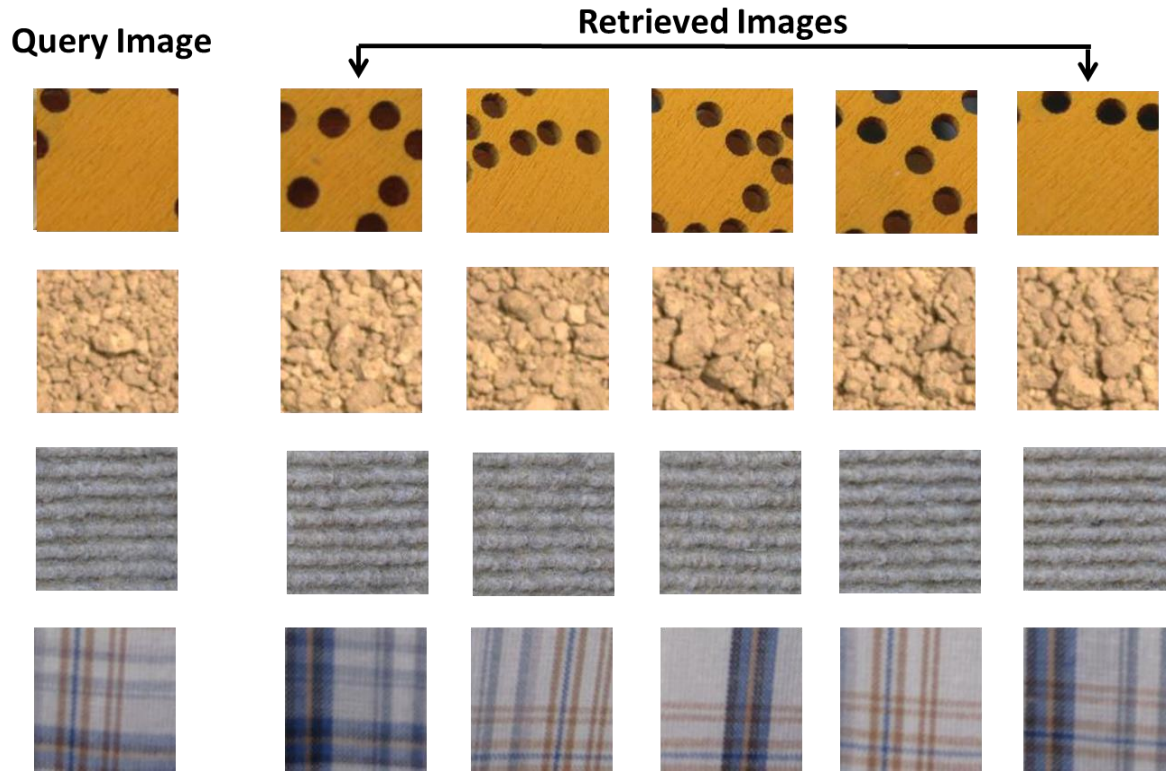

Fig. 12 Query image and retrieved images from Salzburg Texture database.

Table 3: Average Retrieval Rate for different datasets

| Different Feature | Brodatz | OASIS-MRI | ORL | STex |
|---|---|---|---|---|
| CSLBP | 57.08 | 37.25 | 43.45 | 37.14 |
| LEPINV | 58.05 | 37.75 | 27.23 | 36.64 |
| LEPSEG | 65.12 | 38.35 | 37.85 | 27.14 |
| LBP | 70.45 | 44.23 | 41.45 | 52.17 |
| CSLBCoP | 75.45 | 50.45 | 62.15 | 59.14 |
| LTriDP | 77.85 | 50.95 | 55.43 | 57.45 |
| FLNIP(On Single Raw Image) | 79.80 | 53.87 | 63.89 | 61.80 |
| FLNIP(Simple Combination) | 82.32 | 57.85 | 66.14 | 64.43 |
| FLNIP(Weighted Combination using GA) | 84.30 | 58.84 | 68.23 | 67.14 |

**Table 4. Runtime performance and feature length using different methods**

| Different Feature | Feature Length | Feature Extraction Time | Retrieval Time |
|---|---|---|---|
| CSLBP | 16 | 0.0185 | 0.0313 |
| LEPINV | 72 | 0.0709 | 0.0312 |
| LEPSEG | 512 | 0.0360 | 0.0321 |
| LBP | 256 | 0.0181 | 0.0316 |
| CSLBCoP | 1024 | 0.0311 | 0.0325 |
| LTriDP | 768 | 0.0302 | 0.0321 |
| FLNIP(On Single Raw Image) | 256 | 0.0305 | 0.0328 |
| FLNIP(Simple Combination) | 1024 | 0.0312 | 0.0330 |
| FLNIP(Weighted Combination) | 1024 | 0.0356 | 0.0352 |

**Table 5: Comparative study with different distance metrics**

| Distance measure | Brodatz | OASIS-MRI | ORL | STex |
|---|---|---|---|---|
| d1 | 84.54 | 58.84 | 68.23 | 67.14 |
| Euclidean | 73.16 | 53.54 | 58.00 | 56.77 |
| Manhattan | 79.43 | 54.75 | 65.28 | 63.38 |
| Canberra | 70.00 | 55.95 | 65.58 | 52.63 |
| Chi Square | 82.69 | 57.47 | 65.68 | 64.42 |

**Table 6: Comparative study with different standard deviation of gaussian filter**

| Database | $\sigma = X1$ | $\sigma = X2$ | $\sigma = X3$ | d1 |
|---|---|---|---|---|
| Brodatz | 0.4 | 0.7 | 1 | 83.22 |
| | 0.5 | 0.8 | 1 | 84.54 |
| | 0.6 | 0.9 | 1.2 | 82.73 |
| | 0.8 | 1.1 | 1.4 | 80.44 |
| OASIS-MRI | 0.4 | 0.7 | 1 | 57.03 |
| | 0.5 | 0.8 | 1 | 58.84 |
| | 0.6 | 0.9 | 1.2 | 55.38 |
| | 0.8 | 1.1 | 1.4 | 55.08 |
| ORL | 0.4 | 0.7 | 1 | 67.63 |
| | 0.5 | 0.8 | 1 | 68.23 |
| | 0.6 | 0.9 | 1.2 | 65.81 |
| | 0.8 | 1.1 | 1.4 | 64.33 |
| STex | 0.4 | 0.7 | 1 | 66.27 |
| | 0.5 | 0.8 | 1 | 67.14 |
| | 0.6 | 0.9 | 1.2 | 65.79 |
| | 0.8 | 1.1 | 1.4 | 64.91 |

## 4.3 Proposed experimental approach for comparison with deep learning framework

In our best of knowledge, deep learning framework has not been used in these four datasets for the purpose of image retrieval. To compare our model with deep learning framework, we have fine-tuned the widely known VGG 16 and RESNET50 networks pre-trained on ImageNet dataset. Normally, deep learning framework needs huge volume of data, where, Brodatz texture has a total image of 2800, ORL face dataset has a total image of 400 and Salzburg texture has a total image of 7616. To make the datasets appropriate for the deep learning frameworks, we have augmented the datasets applying Random Rotational Augmentation technique. The comparison result is shown in Table 7.

Table 7: Average Retrieval Rate with deep learning frameworks

|  | Brodatz | OASIS-MRI | ORL | STex |
|---|---|---|---|---|
| Fine Tuned VGG 16 | 67.21 | 52.78 | 56.99 | 47.85 |
| Fine Tuned RESNET50 | 55.73 | 48.94 | 55.17 | 33.68 |
| FLNIP(Weighted Combination using GA) | 84.54 | 58.84 | 68.23 | 67.14 |

### 4.3. Error Analysis:

Although our experiments have performed better than others on most datasets, it has not achieved desired results under certain conditions. Since the algorithm has high time complexity owing to genetic algorithm, it is less likely to achieve time efficient results for larger datasets. Moreover, since our method focuses on the multi-resolution approach, it does not produce desired results on datasets having limited variation in textures at different resolutions.

## 5. Conclusion

In this paper, we proposed a novel feature descriptor for content based image retrieval named Fractional Local Neighborhood Intensity Pattern (FLNIP). We have adopted a multi-resolution approach by applying the feature on a set of filtered images obtained by applying gaussian filters on the raw image to find out dominant texture features at different resolutions. This has been followed by using genetic algorithm based learning to perform weighted combination of the different distances obtained between query and database images to reduce the distance between images of similar classes. The novelty in our method lies in the fact that we calculate the sign and magnitude information in one pattern and do not propose separate sign and magnitude components unlike previously developed approaches for image retrieval thus reducing the feature dimension. The method has been evaluated on standard databases and compared with existing techniques by calculating the precision, recall and F-score values for all of them. The feature extraction and image retrieval time is also competitive with most approaches. Thus, the technique can be adapted for efficient image retrieval in real time systems.

**Acknowledgement:** Authors would like to thank Subhramanuim Muarala of Indian Institute of Technology, Ropar for helpful discussion and suggestion to improve the quality of paper.